\documentclass[journal]{IEEEtran}
\ifCLASSINFOpdf
\else
\fi

\usepackage{amsmath}
\usepackage{hyperref}
\usepackage{tcolorbox}
\usepackage{booktabs}
\usepackage{enumitem}
\begin{document}
%
% paper title
% Titles are generally capitalized except for words such as a, an, and, as,
% at, but, by, for, in, nor, of, on, or, the, to and up, which are usually
% not capitalized unless they are the first or last word of the title.
% Linebreaks \\ can be used within to get better formatting as desired.
% Do not put math or special symbols in the title.
\title{Knowledge-intensive Language Understanding for Explainable AI}
%
%
% author names and IEEE memberships
% note positions of commas and nonbreaking spaces ( ~ ) LaTeX will not break
% a structure at a ~ so this keeps an author's name from being broken across
% two lines.
% use \thanks{} to gain access to the first footnote area
% a separate \thanks must be used for each paragraph as LaTeX2e's \thanks
% was not built to handle multiple paragraphs
%

\author{Amit Sheth,\IEEEmembership{}
        Manas Gaur,\IEEEmembership{}
        Kaushik Roy,\IEEEmembership{}
        Keyur Faldu\IEEEmembership{}
        % <-this % stops a space
\thanks{Amit Sheth is the founding director of the Artificial Intelligence Institute at the University of South Carolina (aiisc.ai, \#AIISC), He is a fellow of IEEE, AAAI. AAAS, and ACM, e-mail: amit@sc.edu}
%stops a space
\thanks{Manas Gaur is a Ph.D Student at AIISC focusing on Knowledge-infused Learning, e-mail: mgaur@email.sc.edu}% <-this % stops a space
\thanks{Kaushik Gaur is a Ph.D. student at AIISC focusing on AI algorithms used in health, social media analysis, and recommendation systems, e-mail: kaushikr@email.sc.edu}% <-this % stops a space
\thanks{Keyur Faldu is the Chief Data Scientist at Embible, Inc., India, e-mail: k@embibe.com}% <-this % stops a space
\thanks{Manuscript accepted to IEEE Internet Computing, September/October , 2021}}

% note the % following the last \IEEEmembership and also \thanks - 
% these prevent an unwanted space from occurring between the last author name
% and the end of the author line. i.e., if you had this:
% 
% \author{....lastname \thanks{...} \thanks{...} }
%                     ^------------^------------^----Do not want these spaces!
%
% a space would be appended to the last name and could cause every name on that
% line to be shifted left slightly. This is one of those "LaTeX things". For
% instance, "\textbf{A} \textbf{B}" will typeset as "A B" not "AB". To get
% "AB" then you have to do: "\textbf{A}\textbf{B}"
% \thanks is no different in this regard, so shield the last } of each \thanks
% that ends a line with a % and do not let a space in before the next \thanks.
% Spaces after \IEEEmembership other than the last one are OK (and needed) as
% you are supposed to have spaces between the names. For what it is worth,
% this is a minor point as most people would not even notice if the said evil
% space somehow managed to creep in.

% The paper headers
\markboth{}{}
% The only time the second header will appear is for the odd numbered pages
% after the title page when using the twoside option.
% 
% *** Note that you probably will NOT want to include the author's ***
% *** name in the headers of peer review papers.                   ***
% You can use \ifCLASSOPTIONpeerreview for conditional compilation here if
% you desire.

% If you want to put a publisher's ID mark on the page you can do it like
% this:
%\IEEEpubid{0000--0000/00\$00.00~\copyright~2015 IEEE}
% Remember, if you use this you must call \IEEEpubidadjcol in the second
% column for its text to clear the IEEEpubid mark.

% use for special paper notices
%\IEEEspecialpapernotice{(Invited Paper)}

% make the title area
\maketitle

% As a general rule, do not put math, special symbols or citations
% in the abstract or keywords.
AI systems have seen significant adoption in various domains. At the same time, further adoption in some domains is hindered by inability to fully trust an AI system that it will not harm a human. Besides the concerns for fairness, privacy, transparency, and explainability are key to developing trusts in AI systems. As stated in describing trustworthy AI  (https://www.ibm.com/watson/trustworthy-ai) “Trust comes through understanding. How AI-led decisions are made and what determining factors were included are crucial to understand.” The subarea of explaining AI systems has come to be known as XAI.  Multiple aspects of an AI system can be explained; these include biases that the data might have, lack of data points in a particular region of the example space, fairness of gathering the data, feature importances, etc. However, besides these, it is critical to have human-centered explanations that are directly related to decision-making similar to how a domain expert makes decisions based on "domain knowledge," that also include well-established, peer-validated explicit guidelines. To understand and validate an AI system’s outcomes (such as classification, recommendations, predictions),  that lead to developing trust in the AI system, it is necessary to involve explicit domain knowledge that humans understand and use. Contemporary XAI methods are yet addressed explanations that enable decision-making similar to an expert. Figure one shows the stages of adoption of an AI system into the real world. 

\begin{tcolorbox}
\textbf{Can inclusion of explicit knowledge help XAI provide human-understandable explanations and enable decision making?}
\end{tcolorbox}
\section*{Methods for Explainable AI: Opening the Black Box}
The availability of vast amounts of data and the advent of deep neural network models have accelerated the adoption of AI systems in the real world, owing to their significant success in natural language processing, computer vision, and other data-intensive tasks. However, despite the advances in performance across these tasks, deep learning models remain a black box, i.e., it is extremely hard to understand how the inputs map to the outputs. Recent research in XAI has attempted to address several aspects of "opening this black box" to help humans, both the system users and domain experts, understand such models' functioning and decision-making process \cite{gaur2021semantics}. 
\begin{figure*}[!h]
    \centering
    \includegraphics[width=\textwidth]{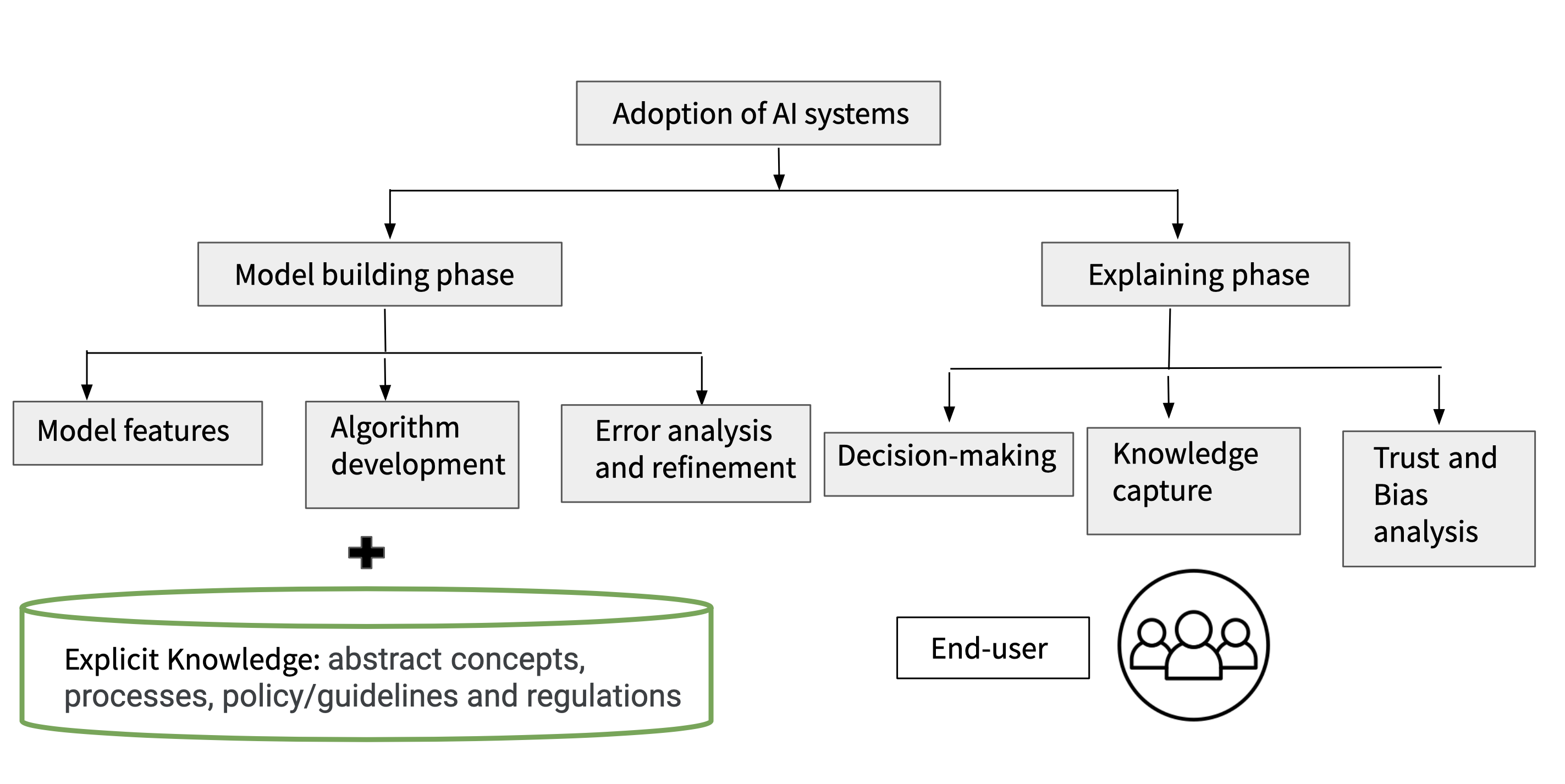}
    \caption{Adoption of AI systems occurs in two stages - the model building phase and the model explanation phase. Explicit knowledge as abstract concepts, processes, policy/guidelines, and regulations are essential to infuse into the AI system for sensible explanations comprehensible to humans.}
    \label{fig:my_label}
\end{figure*}

We now describe and provide references for four main approaches in state-of-the-art XAI for natural language processing that generate explanations from low-level model features in the box below \cite{tjoa2020survey}: 
\begin{tcolorbox}
\scriptsize
\textbf{Approaches and key references for  XAI in natural language understanding}
\begin{itemize}
    \item \textbf{First derivative Saliency based methods} explain the decision of an algorithm by assigning values that reflect the importance of input features in their contribution to that decision in the form of a gradient map:
    \begin{enumerate}
        \item Ribeiro, Marco Tulio, Sameer Singh, and Carlos Guestrin. "" Why should I trust you?" Explaining the predictions of any classifier." Proceedings of the 22nd ACM SIGKDD international conference on knowledge discovery and data mining. 2016.
        \item Zafar, Muhammad Rehman, and Naimul Mefraz Khan. "DLIME: A deterministic local interpretable model-agnostic explanations approach for computer-aided diagnosis systems." arXiv preprint arXiv:1906.10263 (2019).
        \item Lundberg, Scott M., and Su-In Lee. "A unified approach to interpreting model predictions." Proceedings of the 31st international conference on neural information processing systems. 2017.
    \end{enumerate}
    \item \textbf{Layer-wise relevance propagation} decompose the prediction of a deep neural network for a specific example into individual contributions from sub-parts of the text:
    \begin{enumerate}
        \item Montavon, Grégoire, et al. "Layer-wise relevance propagation: an overview." Explainable AI: interpreting, explaining and visualizing deep learning (2019): 193-209.
        \item Yang, Yinchong, et al. "Explaining therapy predictions with layer-wise relevance propagation in neural networks." 2018 IEEE International Conference on Healthcare Informatics (ICHI). IEEE, 2018.
        \item Samek, Wojciech, et al. "Interpreting the predictions of complex ml models by layer-wise relevance propagation." arXiv preprint arXiv:1611.08191 (2016).
    \end{enumerate}
    \item   \textbf{Input perturbations} measure how input changes affect activations and features:
    \begin{enumerate}
        \item Ribeiro, Marco Tulio, Sameer Singh, and Carlos Guestrin. "" Why should I trust you?" Explaining the predictions of any classifier." Proceedings of the 22nd ACM SIGKDD international conference on knowledge discovery and data mining. 2016.
        \item Lundberg, Scott M., and Su-In Lee. "A unified approach to interpreting model predictions." Proceedings of the 31st international conference on neural information processing systems. 2017.
    \end{enumerate}
    \item   \textbf{Attention models} compute focus areas in the text during model decision-making:
    \begin{enumerate}
        \item Bahdanau, Dzmitry, Kyunghyun Cho, and Yoshua Bengio. "Neural machine translation by jointly learning to align and translate." arXiv preprint arXiv:1409.0473 (2014).
        \item Yang, Zichao, et al. "Hierarchical attention networks for document classification." Proceedings of the 2016 conference of the North American chapter of the association for computational linguistics: human language technologies. 2016.
        \item Vaswani, Ashish, et al. "Attention is all you need." Advances in neural information processing systems. 2017.
    \end{enumerate}
\end{itemize}
\end{tcolorbox}
% Note that keywords are not normally used for peerreview papers.
%\begin{IEEEkeywords}
%Knowledge Graphs, Knowledge Infusion, Neuro-Symbolic AI, Explainability, Interpretability, Black-Box Deep Learning, Mental Healthcare, Education Technology
%\end{IEEEkeywords}

% For peer review papers, you can put extra information on the cover
% page as needed:
% \ifCLASSOPTIONpeerreview
% \begin{center} \bfseries EDICS Category: 3-BBND \end{center}
% \fi
%
% For peerreview papers, this IEEEtran command inserts a page break and
% creates the second title. It will be ignored for other modes.
\IEEEpeerreviewmaketitle

\section*{Evaluation of Explanations}
Prior research in assessing the quality of the explanations generated by the XAI system has utilized methods like majority voting over crowdsourcing, visual inspection, annotator agreements, etc. These evaluation metrics are intuitive, but they relegate domain experts to mere annotators of the AI system. Developing a good quality XAI system requires domain experts in the annotation, supervision, and evaluation phases [\cite{roy2021knowledge,roy2021knowledge2,gaur2021can}]. For this purpose, domain experts require explanations that are in the form an expert working in that domain or that application would give, using the language and concepts normally employed by a person working in that field. For example, in the medical domain, the outcome of a model needs to be explained by positioning against conceptual knowledge contained in clinical guidelines. Analysis of word-level and token level features is of little to no use to a domain expert during evaluation \cite{lewis2020retrieval}. 

Evaluation of the quality of explanations using the methods mentioned in the box above requires in-depth knowledge of the mathematical operations such as derivatives, layer-wise feature mapping, perturbations, attention mechanisms, and others. Giplin et al., present a survey of “explaining” explanations and show that human evaluators are necessary to evaluate explanations produced by a model \cite{gilpin2018explaining}. Due to the nature of the explanations, current evaluation of the explanations is limited to analysis of the word and token level feature importances once a suitable visualization mechanism, such as a saliency map, is utilized. Notably, the mathematical expertise required to “open the black box” has been a critical bottleneck in adopting AI systems with explanations. Domain experts require explanations in a language they can easily comprehend or understand to evaluate the system. For example, in the medical domain, the outcome of a model needs to be explained by positioning against conceptual knowledge contained in clinical guidelines. 

Domain-related concepts and clinical guidelines that utilize these concepts to enable outcomes and decisions are stored as explicit knowledge in knowledge graphs (KGs). Thus, methods that incorporate KGs to provide a conceptual level explanation of the model outcome could improve explanations and ease of evaluating AI systems. Popular metrics in language understanding such as, BLEU, ROUGE-L \cite{papineni2002bleu}, QBLEU4 \cite{su2020multi}, BLEURT \cite{sellam2020bleurt} need to be augmented to allow evaluation of the system’s explicit knowledge guided decision-making capabilities. This will lead to trust in the systems by end-users and speedy adoption into the real world. 
\begin{tcolorbox}
\textbf{Explicit knowledge based XAI methods enable trust by explaining the AI systems decision making to the domain expert or end-user in a language and forms  they can easily comprehend.}
\end{tcolorbox}
\section*{Techniques that use explicit knowledge to provide explanations to outcomes}
Recent efforts in the deep learning and NLP community have focused on developing benchmark datasets that would require explicit knowledge \cite{manas2021knowledge}. For instance, consider the task of goal oriented question generation where the goal is to meet the information seeking behavior of the end-user. In such a task, the user provides a query: ``Tell me about the tourism and transportation in Hyderabad”. Leveraging a pre-trained T5 model would generate the following question: “What is tourism” or “What is transportation in Hyderabad”, which are not interesting or relevant to the end-user. In such a task, the end-user is seeking information on tourism and transportation entities within Hyderabad. It is necessary for the deep neural network to develop a good passage retriever and question generator module to obtain contextual questions. This is because the answer lies in a separate but semantically connected passage. Moreover, such a task might require retrieval of a large number of relevant passages as the query is open domain and not factoid. For example, a subsequent question “Tell me about tourism in the Charminar in Hyderabad” will go through the links in Hyderabad, followed by Tourist attractions in Hyderabad, and ending at the article on Charminar (concept flow). This is known as multihop open question answering (ODQA).  A developed retriever/generator pipeline would help end users realize and reason over the questions generated by the model through the support of relevant passages \cite{guu2020realm}. Likewise, the same model could utilize clinical questionnaires with definitions or a clinical manual (such as Diagnostic and Statistical Manual of Mental Disorders (DSM-5) or Structured Clinical Interview for DSM-5) to generate relevant questions \cite{gaur2018let}. Figure 2 shows another example where concept flow occurs in the health-care domain.  The example shows that there is a need for a clear explanation on how the question answering takes place that makes sense to the domain expert. This allows the domain expert to evaluate the questions that lead to trust in the system. 
\begin{figure*}[!h]
    \centering
    \includegraphics[width=\textwidth]{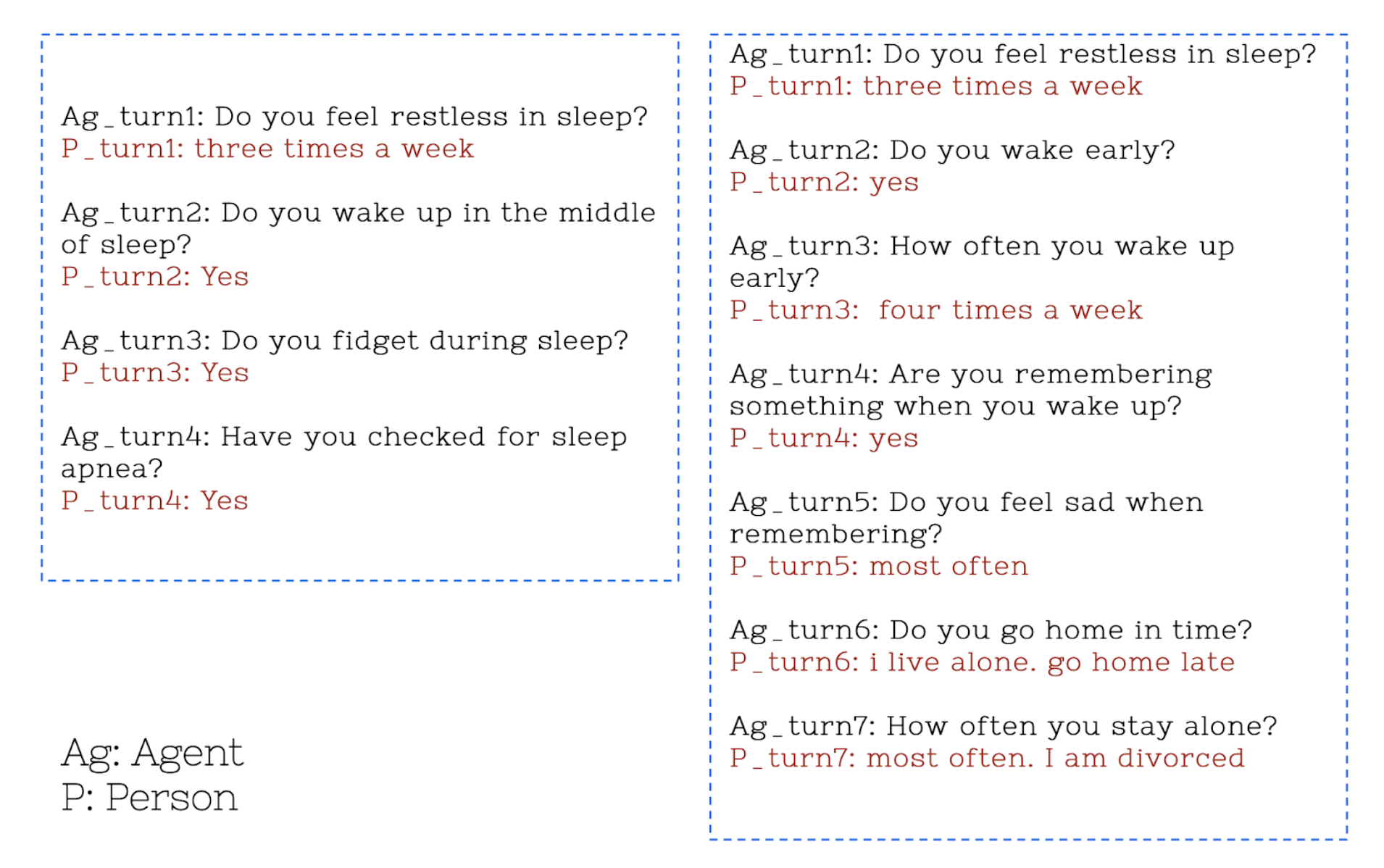}
    \caption{Concept Flow-based Question Generation. Left is generated from a pre-trained T5 fine-tuned for question generation. Right is generated using a T5 fine-tuned on relational context (question and answers) under the supervision of ConceptNet. The difference in the two multi-turn question-answers is that (b) has context-specific questions that drill-down from high-level questions to problem-focused questions. }
    \label{fig:my_label}
\end{figure*}
\section*{From General Language to Knowledge-Intensive Language Understanding}
The NLP community has set up a set of tasks across various benchmark datasets called General Language Understanding Evaluation (GLUE) tasks. They test a variety of natural language tasks such as textual entailment, textual similarity, and duplicate detection. However, recent research has shown that such tasks do not require external knowledge as most tasks are close domain. For example, open domain question answering requires external knowledge to narrow down the scope of passages where the answer lies in. Such tasks are known as Knowledge Intensive Language Understanding (KILU). KILU is a new unified benchmark to help AI researchers build models that are better able to leverage real-world knowledge to accomplish a broad range of tasks. Models that are better able to leverage real-world knowledge do well in these tasks \cite{wang2020k,faldu2021ki}. Traditional explanation methods focus on GLUE tasks. However, since GLUE tasks don’t test if the model can leverage knowledge, the explanations generated are of limited utility to humans. As explained below, this requires abstraction, contextualization, personalization and a variety of knowledge sources to capture information similar to how a human does.
\begin{tcolorbox}
\textbf{To provide explanations to KILU tasks, the model should leverage explicit knowledge and perform abstraction, contextualization, personalization and utilize a variety of knowledge sources.}
\end{tcolorbox}
The use of explicit knowledge in providing explanations achieves the following key capabilities:

\textbf{Abstraction} - The task of mapping low-level features to higher-level human-understandable abstract concepts is known as abstraction. Humans often speak in terms of higher-level abstract concepts when explaining their decision to a user. AI systems also need to explain decisions to the end users using abstract domain-relevant concepts constructed from low-level features and external knowledge in a KG. 

\textbf{Contextualization} - Contextualization is interpreting a concept with reference to relevant use or application. Domain experts contextualize the problem within the domain of a particular disease, for example, depression with its common symptoms and medications. This enables better decision making such as more accurate treatments.

\textbf{Personalization} - Identifying data point-specific information and integrating it with external knowledge to construct a personalized knowledge source is known as personalization. For example, a person's depressive disorder can be due to family issues, relationship issues, and clinical factors. All of these affect the context specific to the individual and consequently affect his symptoms and medications differently than that for another person. 
\begin{figure*}[!h]
    \centering
    \includegraphics[width=\textwidth]{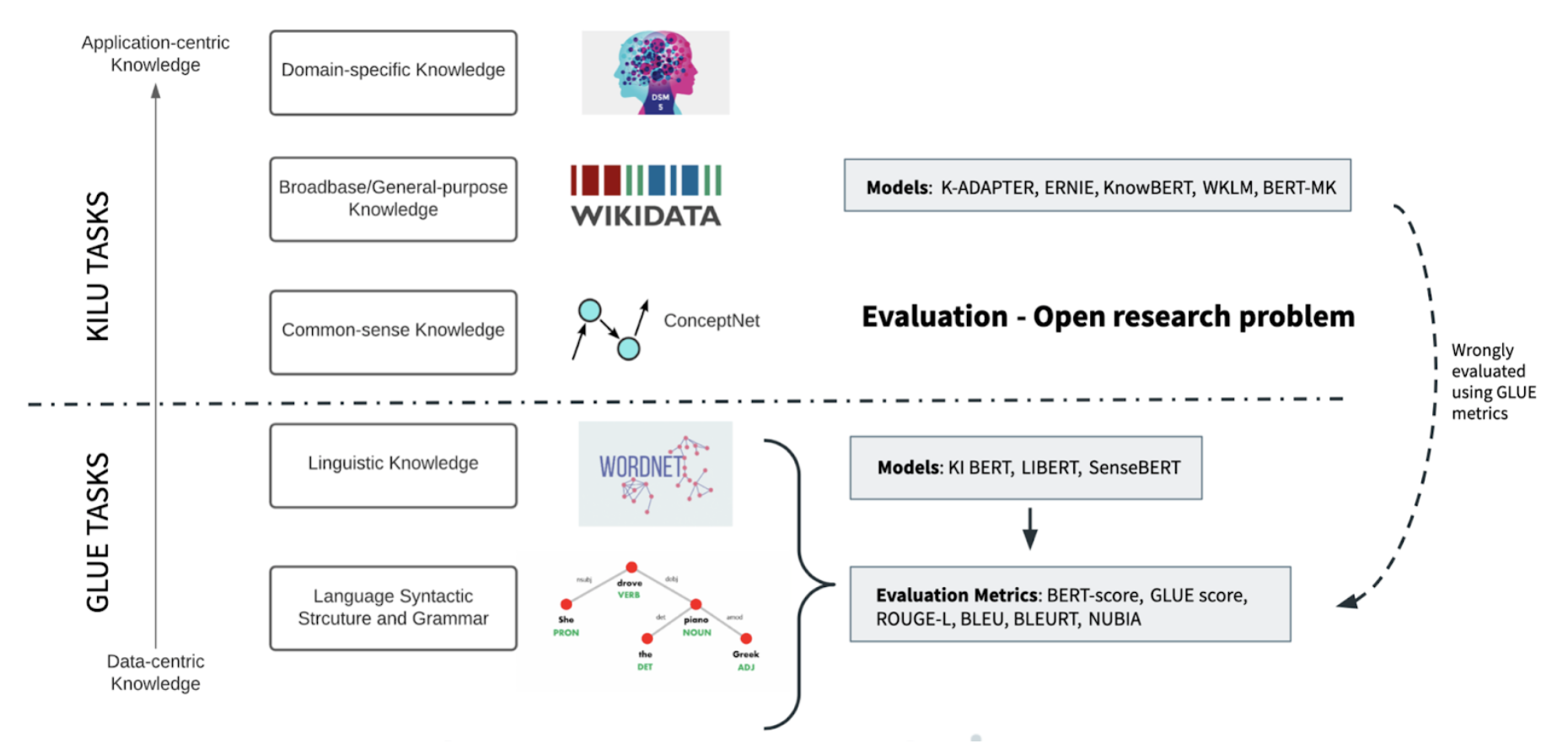}
    \caption{Different sources of knowledge. GLUE tasks are evaluated using BERT-score, GLUE score, ROUGE-L, BLEU, BLEURT, NUBIA metrics \cite{wang2020k}. However, for knowledge intensive knowledge understanding, evaluations require domain knowledge guided explanations.}
    \label{fig:my_label}
\end{figure*}

\textbf{Variety of knowledge capture that humans utilize} - Humans conceptualize by processing information through different levels of knowledge at varying levels of abstractions. Figure 3 shows the different types  of knowledge that humans use including but not limited to syntactic, structural , linguistic, common-sense, general, and domain-specific knowledge. Figure 4 Shows an example of how humans process information by performing personalization through stored historical interactions with the system, contextualization via various sources of knowledge and abstraction through a target source understandable to the end-user. Attempts have been made to infuse knowledge from multiple knowledge graphs to improve domain understanding [\cite{faldu2021ki}]. \begin{figure*}[!h]
    \centering
    \includegraphics[width=\textwidth]{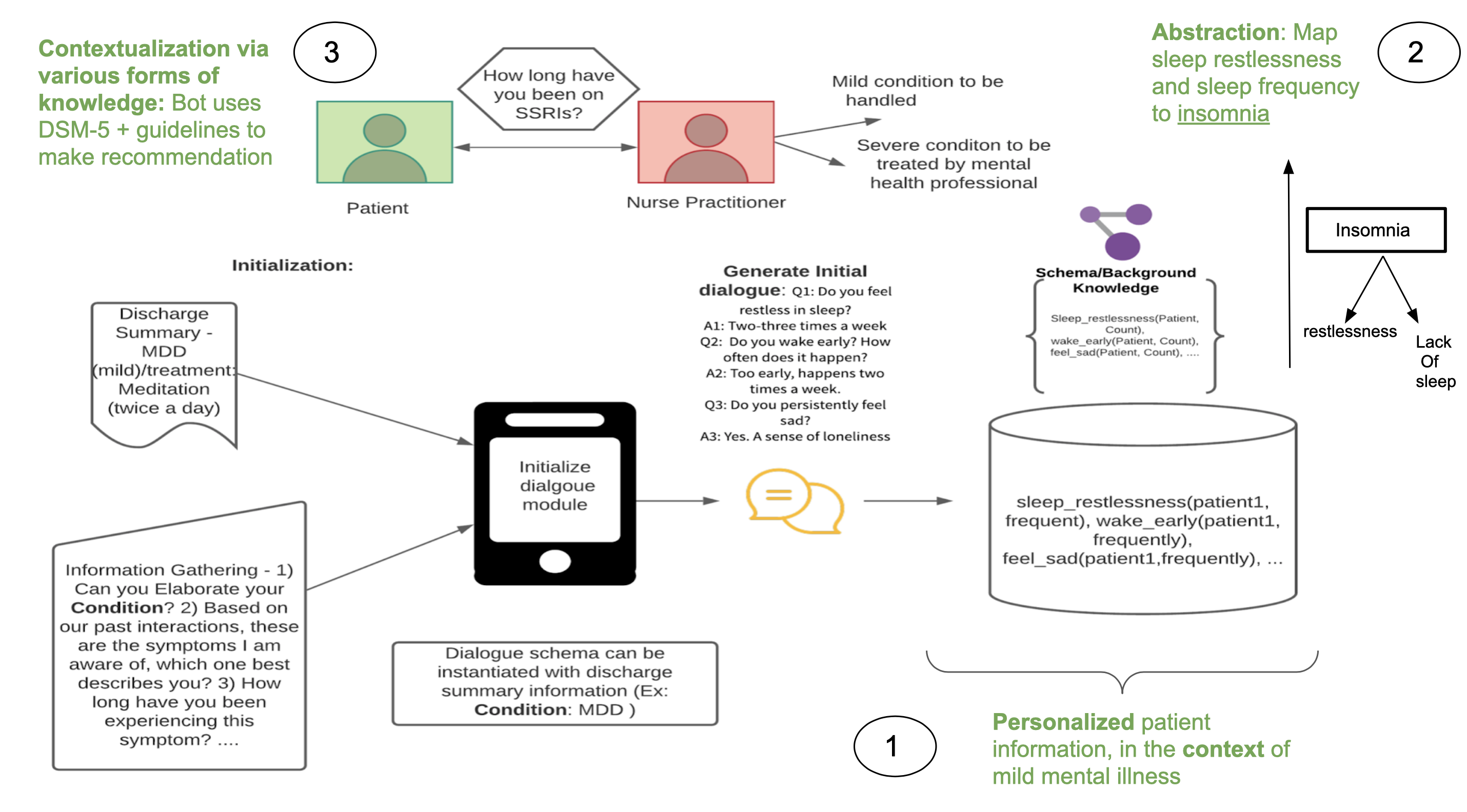}
    \caption{The background knowledge in the form of discharge summaries, transcripts of clinical diagnostic interview, and filled clinical questionnaires (e.g. PHQ-9) created a personalized profile of a user suffering from Major Depressive Disorder (MDD) (\textbf{Step 1}). Domain knowledge abstracts their attributes to reveal insomnia (\textbf{Step 2}). The clinical guidelines (process knowledge) use information about their insomnia to make a recommendation (\textbf{Step 3})}
    \label{fig:my_label}
\end{figure*}  
\section*{Conclusion}
Recent progress in XAI to explain black-box models largely focuses on explanations that map low-level model features to explanations on model decisions or describe the computational paths such as deep network activations that lead to model outcomes. These “system oriented explanations” do little for a domain-expert or an end user who need to be able to trust the AI system’s decision making process, and its adherence to the real-world processes, rules and guidelines. For this, the XAI needs to offer explanations that the end-user or domain expert can easily comprehend. A user does not think in terms of low-level features, nor he seeks to understand the inner workings of an AI system. He thinks in terms of  abstract, conceptual, process-oriented, and task-oriented knowledge external to the AI system. Such external knowledge also needs to be explicit (e.g., as modeled by a knowledge graph), not implicit (i.e., implied by statistics or a vector representation). Recent efforts in knowledge-infused learning \cite{sheth2019shades}, a form of neuro-symbolic AI that utilized explicit external and usually human-curated knowledge, can generate reasonable explanations for users who want to trust an AI system. Thus, explicit external knowledge must be infused into a black-box AI model to generate explanations from low-level features that the domain expert or end-user can understand (see Figure 4). This article also shows the need to develop better natural language understanding benchmarks beyond GLUE that can effectively test the ability of the AI system to explain decisions in a human-understandable manner.
\paragraph*{\textbf{Acknowledgements}}This research is support in part by National Science Foundation (NSF) Award \# 2133842 “EAGER: Advancing Neuro-symbolic AI with Deep Knowledge-infused Learning.” Any opinions, findings, and conclusions or recommendations expressed in this material are those of the author(s) and do not necessarily reflect the views of the NSF.
\bibliographystyle{ieeetr}
\bibliography{references.bib}

% that's all folks
\end{document}